\documentclass[letterpaper, 10 pt, conference]{ieeeconf}  

\IEEEoverridecommandlockouts                              
\usepackage[utf8]{inputenc}
\usepackage[T1]{fontenc}

\title{\LARGE \bf
PAM:Point-wise Attention Module for 6D Object Pose Estimation
}

\author{Myoungha Song$^{1}$, Jeongho Lee$^{2}$, Donghwan Kim$^{3}$
\thanks{$^{1}$,$^{2}$,$^{3}$ are Center for Intelligent \& Interactive Robotics
Korea Institute of Science and Technology}%
}

\usepackage{graphicx}
\usepackage{subcaption}
\usepackage[style=base]{caption}
\begin{document}

\maketitle
\thispagestyle{empty}
\pagestyle{empty}

\begin{abstract}

6D pose estimation refers to object recognition and estimation of 3D rotation and 3D translation. The key technology for estimating 6D pose is to estimate pose by extracting enough features to find pose in any environment. Previous methods utilized depth information in the refinement process or were designed as a heterogeneous architecture for each data space to extract feature. However, these methods are limited in that they cannot extract sufficient feature. Therefore, this paper proposes a Point Attention Module that can efficiently extract powerful feature from RGB-D. In our Module, attention map is formed through a Geometric Attention Path(GAP) and Channel Attention Path(CAP). In GAP, it is designed to pay attention to important information in geometric information, and CAP is designed to pay attention to important information in Channel information. We show that the attention module efficiently creates feature representations without significantly increasing computational complexity. Experimental results show that the proposed method outperforms the existing methods in benchmarks, YCB Video and LineMod. In addition, the attention module was applied to the classification task, and it was confirmed that the performance significantly improved compared to the existing model. 

\end{abstract}

\section{INTRODUCTION}
 
In this paper, we handle the 6d pose estimation of objects. 6D pose estimation refers to the recognition of objects and the estimation of 3D rotations and 3D translation. The 6D pose estimation has been actively researched recently as it has been used in fileds of  autonomous driving \cite{chen2017multi,geiger2012we,xu2018pointfusion}, augmented reality(AR) \cite{marchand2015pose,hinterstoisser2011multimodal}, and robots' manipulation and grasp \cite{collet2011moped,tremblay2018deep,zhu2014single,wang2019densefusion}. This technique is used for object recognition in front of a vehicle in autonomous driving, in AR to find a transformation from a model coordinate system to a camera coordinate system, or to create a gripping point for a robot to grab an object. Therefore, researchs are being conducted to create models that can be applied well in various environments.

However, the 6D object pose estimation has not been able to create a robust feature that responds to problems such as sensor noise, scene clutter, and various shapes and textures, lighting conditions, and occlusion of objects.

\begin{figure}
     \centering
     \begin{subfigure}[b]{0.3\linewidth}
         \centering
         \includegraphics[width=\textwidth]{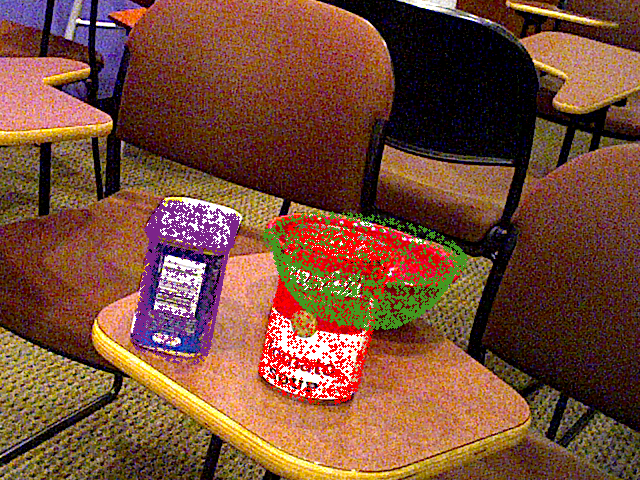}
         \label{pam1}
     \end{subfigure}
     \begin{subfigure}[b]{0.3\linewidth}
         \centering
         \includegraphics[width=\textwidth]{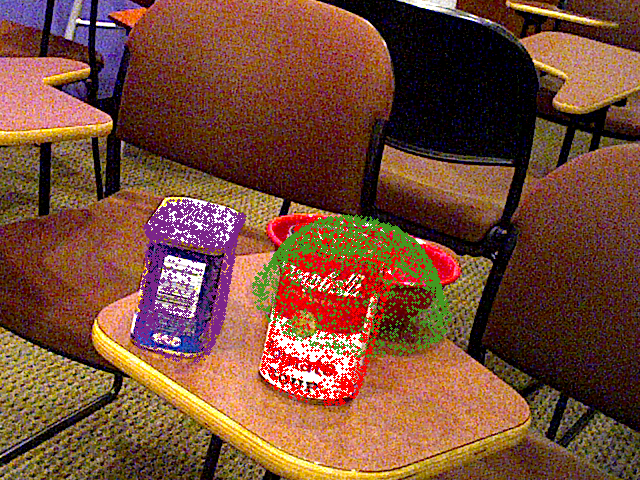}
         \label{origin1}
     \end{subfigure}
     \begin{subfigure}[b]{0.3\linewidth}
         \centering
         \includegraphics[width=\textwidth]{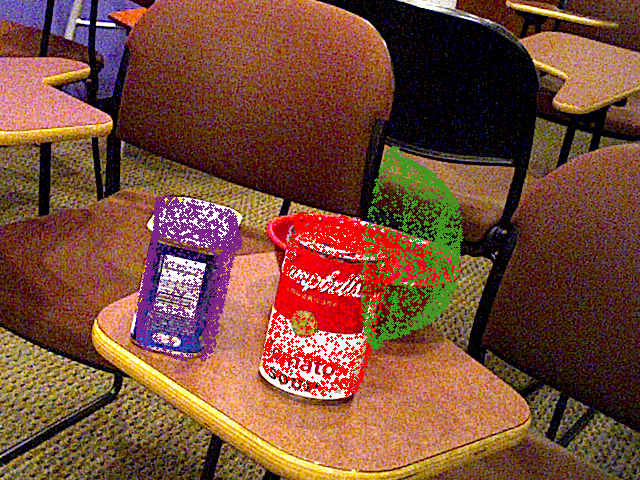}
         \label{posecnn1}
     \end{subfigure}
     \begin{subfigure}[b]{0.3\linewidth}
         \centering
         \includegraphics[width=\textwidth]{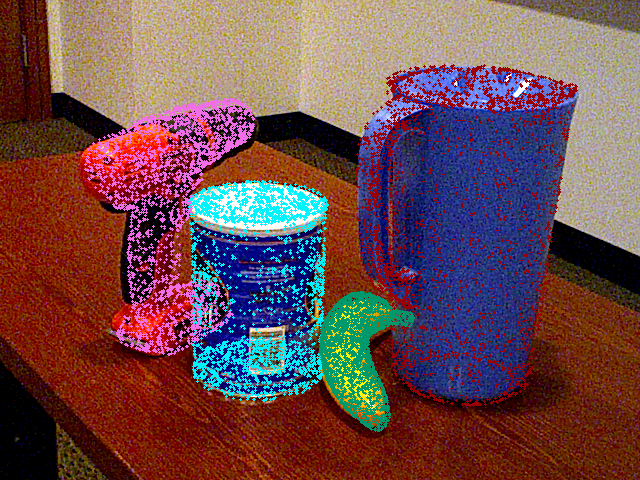}
         \caption{Ours}
         \label{pam2}
     \end{subfigure}
     \begin{subfigure}[b]{0.3\linewidth}
         \centering
         \includegraphics[width=\textwidth]{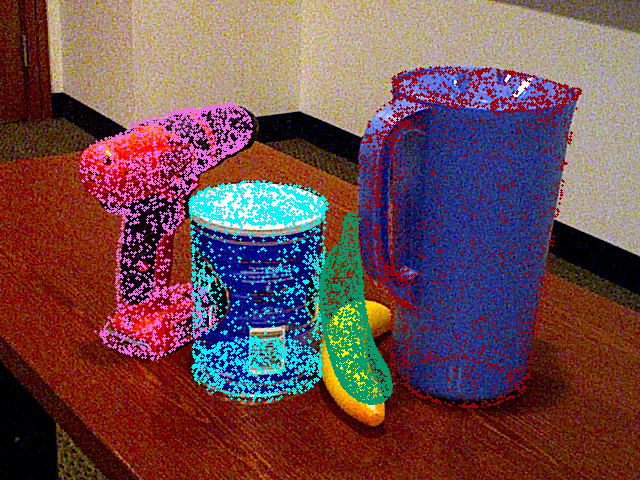}
         \caption{DenseFusion \cite{wang2019densefusion}}
         \label{origin2}
     \end{subfigure}
     \begin{subfigure}[b]{0.3\linewidth}
         \centering
         \includegraphics[width=\textwidth]{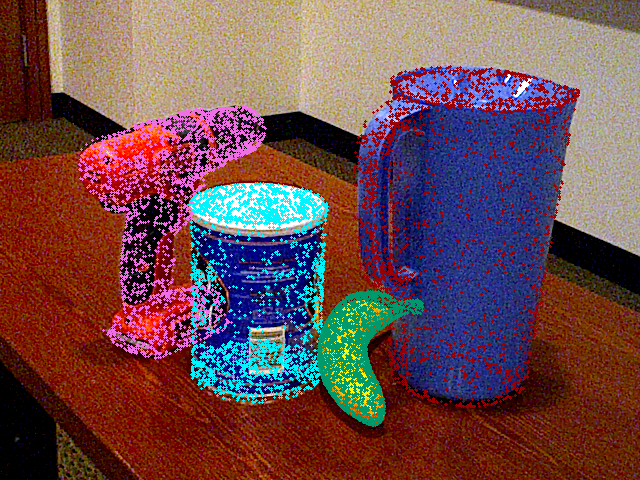}
         \caption{PoseCNN \cite{xiang2017posecnn}}
         \label{posecnn2}
     \end{subfigure}
        \caption{The result is a visualization of three models. (a) is the network to which our point-wise attention module is added, (b) is the original densefusion, and (c) is the result of Posecnn+ICP. Looking at the results, it shows that the pose of (a) is robustly estimated even in the case of occlusion.}
        \label{fig:1}
\end{figure}

To solve these problems, we need to create a robust and discriminative feature representation that will allow us to find a pose from the data.Therefore, in this paper, we aim to estimate the pose as shown in fig.~\ref{fig:1} through robust and discriminative features.

Traditional method created handcraft features in \cite{hinterstoisser2012model,lowe1999object} 2D images and 3D models to estimate posture through the correspondence between 2D and 3D. However, this method was not generalized when there was variability in lighting, or an object's occlusion.

In recent years, deep learning and machine learning technologies have made many advances, and methods of learning with end-to-end have been introduced, and it has become possible to create a stronger feature attention unlike before. In addition, we were able to obtain a generalized model for lighting condition problems and occlusion, which were difficult in traditional method, and the performance was better than before.

In \cite{kendall2017geometric,sundermeyer2018implicit,mousavian20173d}, the feature was extracted from the RGB  using CNN and the pose of the object was estimated directly.

In \cite{rad2017bb8,pavlakos20176,kehl2017ssd,tekin2018real,peng2019pvnet,hu2019segmentation,do2019real} regressed the coordinates of sparse 2D keypoints by learning features from RGB using CNN structure. In \cite{pavlakos20176,oberweger2018making,hu2019segmentation}, a 2D keypoints were obtained on an RGB image through a heatmap representation in as for the methods using keypoints, 6D poses were found through the PnP algorithm for the relationship between 2D-3D using the keypoints obtained by the 2-stage method. These methods have the advantage of speeding up the pose by using only the RGB image and solving the occlusion problem better than the existing methods, but there are limitations in performance than the RGB-D method in that there is no geometric information of the object.

Most of the RGB-D method \cite{xiang2017posecnn,li2018unified,jafari2018ipose} extract features using CNN structure from RGB, estimate the rotation and translation of coarse objects, and then refine the pose through refinement process using ICP(Iterative Closet Point) algorithm \cite{besl1992method} Estimated. Unlike the existing methods, these methods use depth information to estimate a more accurate pose, but the refinement process using ICP takes a very long time and is difficult to apply in real-time.

To solve this problem, \cite{wang2019densefusion,xu2019w,cheng20196d} estimated the pose much more accurately than using only RGB at real-time by designing a network that extracts and fuses features from RGB and Depth's heterogenous data space, respectively, and forms a feature attention. However, these methods do not properly extract local features in the geometrical information.

Recently, researchs were conducted to improve performance by increasing representation power in 2D and 3D-related tasks using the attention technique \cite{cheng20196d,park2018bam,woo2018cbam,li2019zoom,wang2017residual,anderson2018bottom,gehring2016convolutional,bahdanau2014neural,velivckovic2017graph,wang2019graph,poria2017multi,you2018pvnet,sun2020pointgrow,xie2018attentional,hu2018squeeze,vaswani2017attention}. In particular, in the paper of \cite{cheng20196d}, a method was used to estimate poses by creating a feature representation considering the inter-intra correlation between RGB and depth using the attention technique.

In this paper, we propose a network that estimates the pose using Point-wise Module.This network finds the correct pose value by making the representation power of the feature obtained by applying the attention mechanism stronger. In the network, we emphasize the important elements in the feature map extracted point-wise through the PAM. PAM consists of two paths: Channel Attention Path(CAP) and Geometric Attention Path(GAP). CAP uses an inter-channel relationship to create an attention map of what our network wants to emphasize. GAP creates an attention map about where network want to emphasize using local features. Finally, it combines the attention maps from the two paths to find out what features to highlight to find the pose. This method can increase pose accuracy without increasing computational complexity with a small increase in parameters.

To evaluate our method, We experiment with two benchmarks YCB Video dataset\cite{xiang2017posecnn} and LineMod dataset\cite{hinterstoisser2011multimodal} for 6D pose estimation. We get better performance than the previous methods for the two benchmarks. In addition, we experimented with the ModelNet40 dataset to see if our attention structure was applicable to other 3d point cloud tasks. As a result, it was confirmed that our PAM is applicable in classification tasks and performance is improved. To summary, contributions of this work are follows:

\begin{itemize}
\item The proposed PAM enhances the representation power of the feature in a very simple and efficient way by using attention mechanism.
\item We proposed a new structure by combining PAM with the architecture of 6D pose and classification tasks, and this structure can perform tasks more accurately than the existing structure.
\item Our network has now achieved better performance than existing methods on the YCB Video Dataset and Linemod benchmarks. In addition, our PAM was applied to the classification model to evaluate ModelNet40 dataset and shows noticeable performance improvement.

\end{itemize}

This paper is organized as follows.  In Section 2, we explain what studies have been done to solve the pose problem. Section 3, we describe the overall network structure and point-wise attention module. In Section 4, our network finds the optimal hyper parameters and networks through ablation study and shows the results of evaluating the benchmark. the final concluding remarks are given Section5.

\begin{figure*}
    \centering
    \includegraphics[scale=0.65]{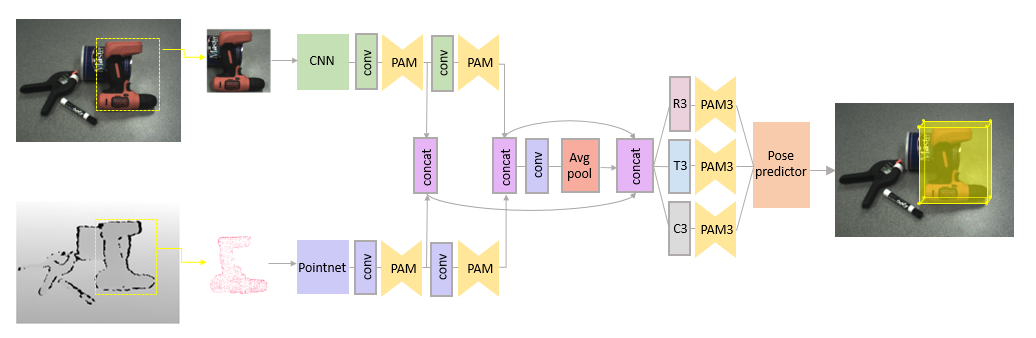}
    \caption{This figure shows our entire network architecture. Our point-wise attention module is included in the part embedding features and the part predicting rotation, translation, and confidence values. In our model, segmentation finds the area of the object we are looking for. Then, feature are extracted using cnn and pointnet structures, and attention modules are added in the middle to extract features.
Subsequently, the extracted feature is made into a global feature through concat, and the pose value is estimated.}
    \label{fig:2}
\end{figure*}
\section{Related Works}

\subsection{Holistic Method}
The holistic methods are a method that creates a feature from the entire image when an input image comes in and directly obtains the 3D rotation and 3D translation of the object. In the classical template based method \cite{huttenlocher1993comparing,gu2010discriminative,hinterstoisser2011gradient,zhu2014single}, each object was configured with a template, and the pose was found by scanning and matching the test. This method has the advantage of being able to respond well to objects without texture, but the limitation was that if the object had an occlusion, its performance was very poor. After the deep learning method came out, it became possible to learn with end-to-end, and a method of creating a feature from an image and finding a pose using a CNN architecture came out. In \cite{xiang2017posecnn}, feature was extracted through CNN architecture and translation and rotation were directly regressed. In particular, when estimating translation and rotation, they set the valance variable so that the two values are predicted well without bias. However, this method had a limitation in that the value of the valance variable had to be set manually. In a recent \cite{li2018unified}, the pose was estimated by constructing a network that decouples the localization, translation, and rotation from the architecture to the branch responsible for each role when an input image is given. However, we extracted features only from the RGB image through CNN architecture and lacked features to find a pose because we did not use depth information to perform three tasks at the same time. Therefore, a method of constructing a feature using information in the depth map came out \cite{li2018unified}. The feature was extracted not only from the RGB image but also from the depth image through the CNN structure, and each extracted feature was fused in a concat manner to find the pose. However, this method has limitations in that it uses feature extracted without considering geometric characteristics.

\subsection{Keypoint-based Method}
Keypoint-based method construct a 2-stage pipeline that finds a 3D keypoints projected on an image using a hand-craft feature or architecture when an input image is given, and solves the relationship between 2d-3d with a PnP algorithm. Finding a keypoints are much easier than finding rotation and translation directly, so much research has been done. The currently keypoint-based methods can be largely divided into three types: Sparse keypoint, heatmap, and densely keypoint.In \cite{tremblay2018deep}, the keypoints of the object were regressed by bringing the network structure that was used in 2D human pose estimation and showed good performance. The sparse keypoints were estimated by constructing a CNN block and stacking N-pieces to create a feature representation. However, the above methods have a limitation that they do not show good performance when the object is occluded. Among the tasks using deep learning, the 2D detection field regressing the coordinates of the bounding box in the image achieved great success, and studies to apply it appeared. In the study of \cite{kehl2017ssd,tekin2018real,do2019real}, a keypoints were obtained by localizing the position of a 3D bound box on a 2D image using a 1-stage detector structure. In particular, in these structures, it is possible to estimate the pose robustly even for low-resolution, in that it uses features of various levels well. However, these methods also have a limitation that they are sensitive when an object has occlusion. The heatmap methods are a method mainly used in human poses. It is a method that sets the largest value among the heatmap values as a keypoint after representing it with a heatmap to find human joints. In \cite{pavlakos20176,oberweger2018making,oberweger2018making}, a heatmap was created in pixel wise using the CNN structure and the region with the largest value was estimated as a keypoint. The heatmap methods respond better to occlusion than the sparse keypoints method in that it uses the surrounding information, but the object is truncated because the size of the heatmap is fixed. \cite{peng2019pvnet} to compensate for the occlusion and truncation, which are the problems covered by the sparse keypoints and heatmap method above, an attempt was made to find the keypoints by representing the direction toward the keypoints in pixel-wise. For each pixel, a direction vectors were directed to the keypoints, and a hypothesis was made for the keypoints to regression the position of the keypoints through voting. However, this method is also stable in that it is a 2-stage process to find keypoints and solve problems between 3D keypoints in the model, but it is a limitation due to information lost in the process of solving the problem using 3D information projected on 2D.
\begin{figure*}
    \centering
    \includegraphics[scale=0.6]{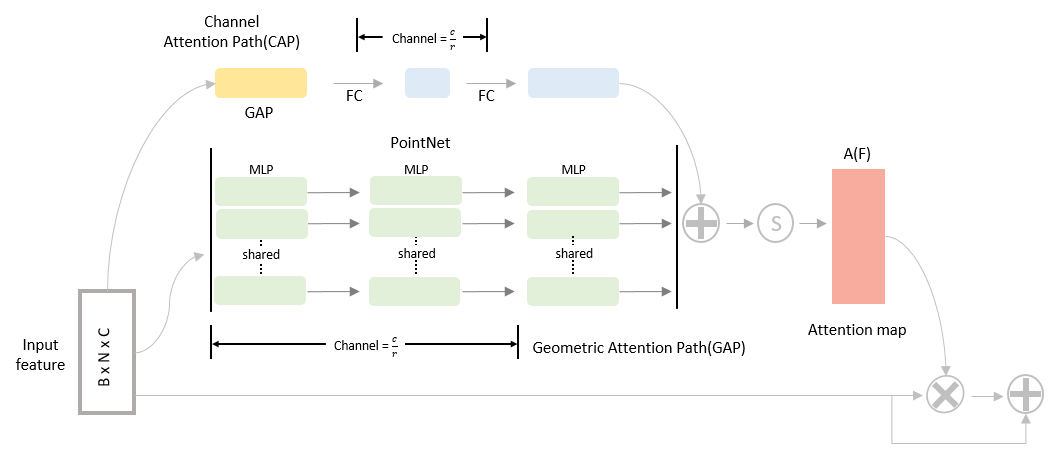}
    \caption{This figure shows our point-wise attention module (PAM). PAM consists of two paths: CAP and GAP. It is made to emphasize what is important in CAP, and is a path made to emphasize where is important in GAP. Each block is composed of convolution 1D, and it prevents computation overhead through reduction ratio. The attention maps created from the two paths are merged through add, and the created attention maps are multiplied with the features of the previous layer, and the values are added back to the feature of the previous layer.}
    \label{fig:3}
\end{figure*}
\subsection{Dense Correspondence Method}
The Dense Correspondence method, which is the opposite method of the keypoint-based method, is a method for estimating pose through feature representation in each pixel or patch unit. In the traditional densely method, when an image is given, it is divided into patch or pixel, and view-point units, and then local feature are extracted by hand-craft from each unit and pose are estimated after that. In \cite{liebelt2008independent,glasner2011aware,sun2010depth}, when an rgb image is given, the image is divided into patch or view-point units, and then the feature is extracted through the feature descriptor to estimate the pose by scoring through each feature through hough-voting. How to do it. Others have attempted to find pose estimation using techniques that are frequently used as machine learning techniques. In \cite{sun2010depth,michel2017global}, random forest was used to predict 3D coordinates for all pixels, and based on that, a hypothesis for pose was generated, and pose was estimated using 2D-3D correspondence. In \cite{kehl2016deep,doumanoglou2016recovering}, the pose was estimated by dividing the image into patch units and extracting feature for each image patch using a CNN encoder and voting. In particular, features extracted using CNN Encoder were combined with random forest, and various information such as rgb, depth, and object rendering were encoded to show good performance. While using depth information together, it provided more and more features and built a robust pose estimation network for occlusion. Since then, the study of dense correspondence has moved to the topic of how to extract RGB and depth information and use it well. In a recent \cite{wang2019densefusion,xu2019w}, to embedding information of different characteristics of the RGB image and depth image, a feature extraction structure suitable for data characteristics was used, and a feature representation that fuses each feature after extracting pixel-wise features Currently achieved SOTA.
\subsection{Attention}
Attention mechanism \cite{vaswani2017attention} is a way for the network to emphasize the influential features. Attention method is used in 2D Image-related Task \cite{li2019zoom,wang2017residual}, language model \cite{anderson2018bottom,gehring2016convolutional}, and 3D-related Task in combination with deep learning. The attention structure used in the language model was applied in combination with CNN. The context information was extracted from the CNN structure and applied to the language structure using the LSTM structure \cite{anderson2018bottom,gehring2016convolutional}. In \cite{hu2018squeeze,wang2017residual}, an influential feature was extracted from the CNN structure and a method was used to pay attention to the feature in the entire structure. Particularly, attention was applied to the relationship between inter channels among features extracted using CNN \cite{hu2018squeeze}. In \cite{wang2017residual}, the CNN module was stacked and a 3D attention map was created to enhance feature. However, \cite{hu2018squeeze} did not have a great effect because only the inter-channel relationship was considered, and in the case of \cite{wang2017residual}, the structure of the entire model became heavy because computation overhead was large in calculating the 3D attention map. 
In \cite{park2018bam,woo2018cbam}, channel attention and spatial attention were respectively composed of structures that complemented the problems of \cite{hu2018squeeze,wang2017residual}, but a reduction ratio was set to prevent computational overhead. Through this structure, \cite{park2018bam,woo2018cbam} showed a significant performance improvement in the classification and detection fields. In \cite{velivckovic2017graph}, the attention method was implemented by graph method for the first time. Method for applying attention to 3D-related tasks began to emerge as the attention manner in the 2D image Task and language model made a good success case. In  \cite{wang2019graph}, the point cloud data was unorder, and due to its irregular characteristics, the convolution kernel could not be applied. Particularly inspired by the attention method of the \cite{velivckovic2017graph}, the attention mechanism was applied as a method of assigning weights when extracting feature for point clouds in a graph method. In \cite{poria2017multi,you2018pvnet}, when fusion of feature from a multimodal dataset, embedding attention was applied to reflect the characteristics of different data spaces to create a more attention aware feature representation. In \cite{sun2020pointgrow}, success was achieved by applying attention to the point cloud generation task. In \cite{xie2018attentional}, the attention method was applied to extract the shape context information of the point cloud more strongly. In \cite{cheng20196d}, the attention mechanism was applied to the 6D pose estimation problem. The correlation and the intra- and inter-modality correlation of RGB and depth information were obtained by constructing an attention map and fusion was used to construct a representation and estimate the pose.

\section{The proposed algorithm }
The 6D object Pose estimation problem is that when an image is given, it finds the object in the scene and estimates the object’s rotation R $\in SO(3)$ and translation $t \in R^3$. More specifically, 6D Pose estimation is finding the rigid transformation $p=[R|t]$ from the object coordinates system of object 3d model we already have to the camera coordinates system of RGB-D transmitted from the camera. The current 6D Object pose estimation task has problems such as lighting condition, sensor noise, occlusion, variation of texture and shape.To solve the problem, it is important to efficiently extract features from RGB image and essential feature from depth image to create robust and discriminative feature representation. But it is hard to extract important feature because rgb and depth are divided into different data spaces. rgb image contains context information, and depth image contains geometrical information. Therefore, it is key to efficiently and well extract each feature from data with different characteristics through the network to increase the representation power. In our paper, we apply attention mechanism to enhance the stronger and more influential values in features extracted from each heterogeneous space, thereby increasing the representation power of features.

\subsection{Overview} 
fig.~\ref{fig:2} shows the overall shape of our structure. In 1-stage, semantic segmentation is used to obtain an object region. In 2-stage, feature extraction and point-wise feature fusion are performed using self-attention. In this process, CAP and GAP are constructed from the extracted features using the PointNet structure to extract more influential feature from each path. Finally, after estimating poses, additionally, iterative refinement algorithm proposed by densefusion network is used to estimate more sophisticated and accurate poses. Section 3.2 describes a semantic segmentation part for extracting a region of an object. At Section3.3, attention mechanism and PAM structure will be described. In Sec3.4, the process of finding the final pose and the iterative refinement process are explained. Finally, in Sec3.5, the designed Loss function is described.

\subsection{Semantic Segmentation}
The first step is localize where the object of interest is located through semantic segmentation in a given RGB image. Through localization, the area of the object was designed from RGB and Depth image to be less affected by other objects or backgrounds. The Encoder section was experimented with a much-used backbone CNN feature extractor structure, and the Decoder used same structure as \cite{wang2019densefusion,xu2019w,cheng20196d} to create N+1 semantic map. What N+1 means is N-th object and Background, and we used the same semantic segmentation result provided \cite{xiang2017posecnn}, focusing only on Pose performance.

\subsection{Feature Extraction and Point-Wise Feature Attention} 
The second step is to independently extract features from RGB and Depth to optimize the pose. to extract the spatial feature from the rgb image, we used the CNN structure to extract the feature, and we extracted it through PointNet structure to extract geometric information from Point cloud. Before each independently extracted feature was fusion, PAM was added to enhance the influential feature value and a representation that fuses the feature.

\subsubsection{Attention Mechanism}
attention mechanism means that the network itself is designed to find influential elements within the feature and enhance the feature. PAM predicts A(F) when feature F from the previous layer comes. Here, A(F) is the same as (\ref{equation:2}) below and consists of the sum of two paths. In (\ref{equation:1}), F applies the Attention to this feature  when a feature is passed from the previous layer and adds it through the product with the previous layer. Through the process of A(F), the influential part of the previous layer is generated as an attention map, and the influential feature is selected by multiplying with the feature. Then, by adding it to the feature again, the influential feature is made stronger. Using Attention, in the process of learning the network, the feature map is well found through the attention map, and the feature representation is created in the form of stronger overall important features.

\begin{equation}
F' = F + F \bigotimes A(F) 
\label{equation:1}
\end{equation}

\begin{equation}
A(F)= \sigma(A_C(F) + A_G(F))
\label{equation:2}
\end{equation}

\subsubsection{Channel Attention Path(CAP)}
The Channel Attention Path is designed to embed information between inter channels. Since the channel value of the feature has a pattern encoding value, it can be said that the process of considering relations between inter channels is important. Therefore, in the process of learning the network, the feature was extracted for the relation between inter channels within the feature, and it was designed to emphasize important features in the channel through this. To implement this process, the Channel Attention branch used in BAM\cite{park2018bam} was changed and applied. In fig.~\ref{fig:3}, the Channel Attention Path makes the feature transferred from the previous layer a vector using global average pooling. This vector encodes global information for each channel. In order to estimate the attention map for all channels, the next layer and shared features were used by using the MLP layer. In this process, a reduction ratio was established to prevent the computation overhead. MLP is reduced to c by reducing the value of c/r to c. The shape is then fixed to match the output and the value from the geometric attention path.
\subsubsection{Geometric Attention Path(GAP)}
In the Geometric Attention Path, it is designed to embed geometric information. Geometric information is information that contains how an object is positioned in 3d, and is very essential for estimating a pose. In the existing BAM\cite{park2018bam} structure, feature are extracted using dilated convolution to contain spatial information of 2D images. However, 2d convolution is not suitable for handling point cloud data, and 3D convolution requires a lot of computation. Therefore, it embedding a local feature well for point cloud data, shows good performance in various 3D-related tasks, and applied it by removing the pooling layer from the widely used pointnet structure. It is designed to reinforce the local feature from the feature passed from the previous layer by embedding the geometric feature well using the Geometric Attention path. In fig.~\ref{fig:3}, geometric information is embedding through the Pointnet structure from feature from the previous layer. In this process, reduction ratio was configured in the layer in front of the pointnet to prevent the computation overhead. 
\subsubsection{Combining Two Attention Paths}
The attention map formed in each path is combined to make the final $A(F)$ by combining the attention map formed in CAP ($A_C(F)$) and GAP ($A_G(F)$). Because the attention maps formed from the two paths have different sizes, use the tensor broacasting to match the size equally. Attention maps of the same size are combined through element-wise summation, and values are made between 0 and 1 using the sigmoid function. The attention feature is created by multiplying the final attention map created in this way by element wise on the input layer, and then adding values to the previous layer to strengthen the channels and geometrically meaningful values.

\subsection{6D object pose estimation with iterative refinement}
The subsequent process estimates the values densely using the point-wise attention embedding feature created through the previous process. The values predicted through the network are designed to predict rotation, translation, and confidences. Here, the confidences is the value that represents the probability of the most accurate pose from the results estimated by point-wise. As in \cite{wang2019densefusion}, the confidences predicted by self-supervised learning can be found to be the most accurate pose value. 

Iterative Refinement is the process of changing the predicted pose from PoseNet to the initial pose to find the pose in detail to the ground truth pose. The RGB features already extracted through posenet are reused in the iterative refinement process, and the feature extract and feature fusion process are performed in the same process as PoseNet using the transformed point cloud values. Predict the rotation and translation through the generated feature and predict the remaining pose  from the initial pose. This process is the same as (\ref{equation:3}), the more repeat it, the more accurate the pose can be found. Repeat K times to converge to the final pose.

\begin{equation}
\tilde{p} = [R_k |T_k ]\cdot[R_{k-1} |T_{k-1}]\cdot[R_{k-2} |T_{k-2}]\cdots[R_0 |T_0]
\label{equation:3}
\end{equation}

\subsection{Loss Function}
The Loss uses the ADD and ADD-S Loss functions proposed in the \cite{xiang2017posecnn} paper, which is often used in the field of 6D pose estimation. The ADD loss function (\ref{equation:4}) uses the sampled point and ground truth point to transform through rotation and translation obtained through the network, and progresses the learning in the direction of minimizing the average value of the distance of the ground truth point.
\begin{equation}
L^{p}_{i} =\frac{1}{M}\sum_{j}||(Rx_j+t) -(\tilde{R}_ix_j+\tilde{t}_i) ||
\label{equation:4}
\end{equation}

Where $x_j$ means a point in the j-th 3D model among randomly selected M pieces. $p=[R|t]$ is the ground truth pose value, and $\tilde{p} = [\tilde{R}|\tilde{t}] $ is the predicted pose value through the network.
In the case of this loss function, as mentioned in the \cite{xiang2017posecnn} paper, there is a very weak part for symmetric objects, so we use ADD-S loss function to learn about symmetric objects. ADD-S(4) loss function is as below (\ref{equation:5}).

\begin{equation}
L^{p}_{i}=\frac{1}{M}\sum_{j}\min_{0<k<M}||(Rx_j+t) -(\tilde{R}_ix_k+\tilde{t}_i) ||
\label{equation:5}
\end{equation}
We have to learn to select a value that is likely to be the correct answer pose from the predicted rotation and translation using the confidence value as applied in the \cite{wang2019densefusion,xu2019w} paper. To do so, multiply the ADD(S) loss function obtained above by the confidence value as suggested in \cite{wang2019densefusion,xu2019w}. After that, confidence was used as a regularization term by applying a log. The equation is as (\ref{equation:6}) below.

\begin{equation}
L =\frac{1}{N}\sum_{i}(L^{p}_{i}c_i -w\log(c_i))
\label{equation:6}
\end{equation}

\begin{table*}[ht]
\centering
\caption{This table was created to compare our model's results to the YCB video dataset with the latest models. The table compares ADD(s)<2cm and AUC values, and shows the performance without refinement on the left side and the result of refinement on the right side.}
\begin{tabular}{|c|c|c|c|c|c|c|c|c|c|c|c|c|c|c|}
\hline
&\multicolumn{6}{|c|}{without Refinement}& \multicolumn{8}{|c|}{with Refinement}\\ 
\hline
 & \multicolumn{2}{|c|}{DenseFusion} &\multicolumn{2}{|c|}{W-PoseNet} &\multicolumn{2}{|c|}{ Our} &\multicolumn{2}{|c|}{ PoseCNN+ICP}  &\multicolumn{2}{|c|}{ DenseFusion}  & \multicolumn{2}{|c|}{ W-PoseNet} &\multicolumn{2}{|c|}{ Our}\\
 \hline
 & AUC & <2cm & AUC & <2cm & AUC & <2cm & AUC & <2cm & AUC & <2cm& AUC & <2cm& AUC & <2cm\\
 \hline
 002 master chel can& 95.2 & 100.0 & 69.2 & 65.9 & 95.4 & 100.0 &68.1 &51.1 &96.4&100.0&72.0&68.6&95.2&100.0\\
 003 cracker box& 92.5 & 99.3 & 87.8 & 90.0 & 93.8 & 99.0 &83.4 &73.3 &95.5&99.5&91.3&93.7&96.2&99.7\\ 
 004 sugar box& 95.1 & 100.0 & 91.5 & 98.5 & 95.8 & 100.0 &97.5 &99.5 &97.5&100.0&95.1&99.8&97.63&100.0\\
 005 tomato soup can& 93.7 & 96.9 & 87.4 & 84.2 & 94.0 & 96.9 &81.8 &76.6 &94.6&96.9&88.9&84.6&94.5&96.9\\
 006 mustard bottle& 95.9 & 100.0 & 93.4 & 100.0 & 96.6 & 100.0 &98.0 &98.6 &97.2&100.0&96.5&100.0&97.7&100.0\\
 007 tuna fish can& 94.9& 100.0& 77.0 & 55.9 & 95.3 & 100.0 &83.9 &72.1 &96.6&100.0&78.8&61.4&96.8&100.0\\
 008 pudding box& 94.7 & 100.0 & 91.8 & 98.81 & 94.8 & 99.5 &96.6 &100 &96.5&100&94.5&100.0&96.1&100\\
 009 gelatin box& 95.8 & 100.0 & 94.6 & 100.0 & 97.7 &100.0&98.1 &100.0 &98.1&100.0&96.0&100.0&98.4&100.0\\
 010 putted meat can& 90.1 & 93.1 & 79.0 & 77.8 & 90.0 & 93.0 &83.5 &77.9 &91.3&93.1&82.6&80.4&91.4&93.1\\
 011 banana& 91.5 & 93.9 & 87.9 & 87.3 & 93.6 & 97.1 &91.9 &88.1 &96.6&100.0&92.8&98.9&96.8&100.0\\
 019 pitcher base& 94.6 & 100.0 & 92.0 & 100.0 & 94.0 & 100.0 &96.9 &97.7 &97.1&100.0&95.0&100.0&97.29&100.0\\
 021 bleach cleanser& 94.3 & 99.8 & 85.2 & 77.4 & 95.2 & 99.8 &92.5 &92.7 &95.8&100.0&89.5&89.0&95.5&100.0\\
 024 bowl& 86.6 & 69.5 & 86.2 & 49.8 & 86.6 & 96.8 &81.0 &54.9 &88.2&98.8&87.7&93.6&87.7&99.3\\
 025 mug& 95.5 & 100 & 84.9 & 79.7 & 95.4 & 100.0 &81.1 &55.2 &97.1&100.0&88.2&90.3&96.8&100.0\\
 035 power drill& 92.4 & 97.1 & 91.1 & 98.8 & 93.1 & 97.1 &97.7 &99.2 &96.0&98.7&93.6&99.5&95.9&98.8\\
 036 wood bolck& 85.5 & 93.4 & 86.3 & 96.3 & 88.9 & 98.4 &87.6 &80.2 &89.7&94.6&87.0&97.5&92.3&99.2\\
 037 scissors& 96.4 & 100.0 & 91.5 & 99.5 & 91.4 & 96.7 &78.4 &49.2 &95.2&100&90.7&97.8&93.4&99.5\\
 040 large marker& 94.7 & 99.2 & 90.9 & 96.3 & 95.6 & 99.9 &85.3 &87.2 &97.5&100.0&92.5&99.4&97.6&100.0\\
 051 large clamp& 71.6 & 78.5 & 71.4 & 74.0 & 70.5 & 78.4 &75.2 &74.9 &72.9&79.2&70.8&79.2&73.0&78.5\\
 052 extra large clamp& 69.0 & 69.5 & 68.0 & 60.4 & 73.9 & 71.7 &64.4 &48.4 &69.8&76.3&69.6&72.7&75.1&76.3\\
 061 foam brick& 92.4 & 100.0 & 92.5 & 100.0 & 92.1 & 100.0 &97.2 &100.0 &92.5&92.5&92.9&100.0&96.0&100.0\\
\hline
MEAN& 91.2 & 95.3 & 85.7 & 85.2 & 91.8 & 96.3 &86.6 &77.9 &93.1&96.8&87.9&90.8&93.4&\textbf{96.9}\\
\hline
\end{tabular}

\label{table:3}
\end{table*}

\section{Experiments}

\subsection{Datasets}
The dataset used the benchmarks, YCB Video dataset\cite{chen2017multi} and Linemod dataset\cite{hinterstoisser2012model} to evaluate the same as the current methods.
\subsubsection{YCB Video Dataset}
YCB Video Dataset is a dataset consisting of objects with various shapes and textures of a total of 21 objects. It consists of a total of 92 video datasets, and the data that combines 80,000 synthetic datasets with 16,189 frames of 80 video datasets as set in the \cite{wang2019densefusion,xu2019w,cheng20196d} papers is a training dataset. The frame is used as a test dataset. \cite{wang2019densefusion,xu2019w,cheng20196d} To evaluate under the same conditions used in the paper, the result of segmentation uses the result of segmentation evaluated in posecnn\cite{xiang2017posecnn}, and the number of iterations is composed of two.
\subsubsection{LineMod}
The Linemod dataset is a dataset consisting of a total of 13 low texture objects. A total of 15,783 frames, 13 objects are made to form different clutter environments and occlusions. In the same way as we did in the \cite{wang2019densefusion,xu2019w,cheng20196d}, 85\% consisted of training datasets, and the remaining 15\% consisted of test datasets.
\subsection{Metric}
We applied the AUC of ADD(S) and ADD(S)<2cm evaluation indicators to the YCB Video Dataset in the same way as we evaluated in the recent \cite{wang2019densefusion,xu2019w,cheng20196d}. In the case of asymmetric objects, we transformed the 3D model we had using the ground truth pose and the predicted pose, and then calculated the difference between the two by calculating the ADD (average distance). In the case of a symmetric object, the distance between points was measured using the minimum value using two pose.
\subsubsection{AUC}
AUC is a measure designed to prevent the wrong measurement of the pose measurement using a fixed threshold as the metric proposed by \cite{xiang2017posecnn}. The use of these evaluation indicators allows the accuracy to be determined by a curve according to the threshold, and the area below the curve can be used for use as a pose evaluation. Therefore, we set the threshold from 0 to 10cm to examine the AUC area.

\subsubsection{ADD(S)<2cm}
ADD(S)<2cm metric sets the minimum allowable error in robot manipulation as 2cm as a threshold, and assumes that the pose is found correctly when the error is smaller than this. Here, ADD(S) was based on ADD for asymmetric objects, and ADD(S) was used for symmetric objects.
\subsection{Implementation Detail}
This paper was conducted on CPU i9-9900X and NVIDIA TITAN Xp 2 GPUs. The CNN structure consists of an encoder and a decoder. In the encoder part, we experimented with resnet18, which extracts context information well, and in the decoder part, the structure used by PsPNet was constructed with 4 up-samples. Three convolutions were used in the geometric attention path, and the attention module was used by experimentally adjusting the number according to the DB. In the Attention module, the reduction ratio was set to 16, which is empirically most optimal through experiments, to minimize computational complexity. The iterative refinement process was set up and tested twice, and the maximum value of the results of more than 20 tests was selected for all experiments.

\subsection{Ablation Study}
\subsubsection{Reduction Ratio}
The advantage of using PAM is that it can improve the performance by increasing the computational complexity of a small amount. The reduction ratio prevents calculation overhead by controlling the number of channels in the module. In Table?, the efficiency of calculation is confirmed by looking at the increase and performance of the parameter according to the change of the reduction ratio. We could confirm that the parameter increase amount decreases as the reduction ratio increases. In TABLE~\ref{table:2}, the performance when the reduction ratio is 16 is higher than that of 4 and 8. In addition, it can be seen that the performance is improved by 3.8\% with a small parameter increase of about 0.35M. We think that overfitting is the reason why the performance is worse when it is 4, 8, and in 32 and 64, the amount of the parameter is too small. Therefore, in the following experiment, the reduction ratio was fixed at 16 to conduct the experiment.

\begin{table}
    \caption{ this table shows the change in parameter increase and performance according to the reduction ratio. As can be seen in the table, if the reduction ratio is too small, the amount of model weight increases and the performance improves, whereas the smaller the ratio, the model weight decreases and the increase in performance decreases.}
    \centering
    \begin{tabular}{|c||c||c||c|}
        \hline
        & Value & Parameters & ADD(S)\\ \cline{2-4}
        \hline
        & base & 21.44M & 94.3\\
        \hline\hline
        & 4 & 23.01M & 97.0  \\ 
        & 8 & 22.17M & 97.2 \\ 
        Reduction Ratio & 16 & 21.79M & \textbf{97.8} \\ 
        & 32 & 21.62M & 97.1 \\ 
        & 64 & 21.53M & 97.3 \\ 
        \hline
    \end{tabular}
    \label{table:1}
\end{table}

\subsubsection{Channel and Geometric Path}

In this experiment, we conducted an experiment to confirm the effect of each attention path we designed. We first experimented with adding both CAP and GAP and a network with nothing. As shown in TABLE~\ref{table:3}, when CAP and GAP were used together, the performance increased significantly from 94.3\% to 97.8\%. Performance increased by 3.3\% from 94.3\% to 97.6\% when using CAP alone, and increased by 2\% from 94.3\% to 96.3\% when using GAP. It can be seen from the results that the two paths each play an important role in improving performance. It follows a similar aspect to the human visual system, and when a person looks at a 3d object, which part to look further (GAP), what to look at (CAP), and the two paths combine to form visual information\cite{park2018bam}.

\begin{table}
    \centering
    \caption{ This table shows the performance change according to the presence or absence of each path. When the channel attention path was added, it showed a large performance increase, and when the geometric path was added together, it showed the largest performance increase.}
    \begin{tabular}{|c||c||c|}
    \hline
    & Reduction Ratio & ADD(S)\\
    \hline
    base & X & 94.3\\
    \hline\hline
    CAP &  & 97.6  \\ 
    GAP & 16  & 96.3 \\
    CAP + GAP &   & \textbf{97.8} \\ 
    \hline
    \end{tabular}
    \label{table:2}
\end{table}

\begin{table*}[ht]
\centering
\caption{This table was created to compare our model's evaluation of the LineMod dataset with other latest methods. The table compares ADD\cite{hinterstoisser2012model}, and shows the performance without refinement on the left side and the result of refinement on the right side. Results are report in units of \%}
\begin{tabular}{|c|c|c|c|c|c|c|c|c|c|c|}
\hline
&\multicolumn{3}{|c|}{without Refinement}&\multicolumn{6}{|c|}{with Refinement}\\ 
\hline
 & DenseFusion &W-PoseNet &Our &Implicit+ICP&SSD-6D+ICP&DenseFusion&DPOD&W-PoseNet &Our\\
 \hline
 ape & 79.5& 86.3& 81.0&20.6&65.0&92.3&87.73&92.8&95.4\\
 bench vi. &84.2&97.4&94.0&64.3&80.0&93.2&98.45&99.6&96.9\\
 camera &76.5&98.3&86.2&63.2&78.0&94.4&96.07&99.0&98.2\\
 can &86.6&97.5&91.5&76.1&86.0&93.1&99.71&99.3&96.9\\
 cat &88.8&97.3&93.0&72.0&70.0&96.5&94.71&99.0&98.0\\
 driller &77.7&95.9&89.4&41.6&73.0&87.0&98.8&97.8&95.3\\
 duck &76.3&93.3&89.0&32.4&66.0&92.3&86.29&96.2&96.8\\
 eggbox &99.9&99.9&99.9&98.6&100.0&99.8&99.91&99.9&99.9\\
 glue &99.4&99.8&99.8&96.4&100.0&100.0&96.82&99.9&99.7\\
 hole p. &79.0&95.7&90.6&49.9&49.0&92.1&86.87&97.3&96.0\\
 iron &92.1&96.6&96.5&63.1&78.0&97.0&100.0&98.6&99.2\\
 lamp &92.3&99.2&94.0&91.7&73.0&95.3&96.84&99.8&98.3\\
 phone &88.0&96.4&94.5&71.0&79.0&92.8&94.69&98.3&97.4\\
 \hline
MEAN & 86.2& 96.4& 92.2&64.7&77.0&94.3&95.15&\textbf{98.2}&97.5\\
\hline
\end{tabular}

\label{table:4}
\end{table*}

\subsection{Evaluation on YCB Video Dataset}
TABLE~\ref{table:3} shows the results of comparing the evaluation of 21 objects of the lastest methods and YCB Video Dataset. It can be seen that the results when the refinement was not performed and when the refinement was performed performed better than other methods. In particular, the performance of ADD(S)<2cm without refinement has risen to a level similar to that of \cite{wang2019densefusion} refinement. It can be seen that AUC performance has been greatly improved in objects such as bowls, bananas, and wood blocks. Objects are objects that are severely occluded by other objects on the test data, and the feature representation through our attention can be said to be more robust to occlusion than existing methods. In addition, although large clamps or extra large clamps are objects that have difficulty in finding poses with symmetric objects, our feature representations that emphasize geometric characteristics more accurately help us find the poses of objects. Looking at the visualized results in fig.~\ref{fig:5} , it was possible to find the correct pose in the case of an occluded bowl or banana that was not found in the existing network. Also, in the case of scissors with half of the object obscured, it was possible to find the correct pose.

\subsection{Evaluation on LineMod Dataset}
\begin{figure*}[htp]
    \centering
    \includegraphics[scale=0.8]{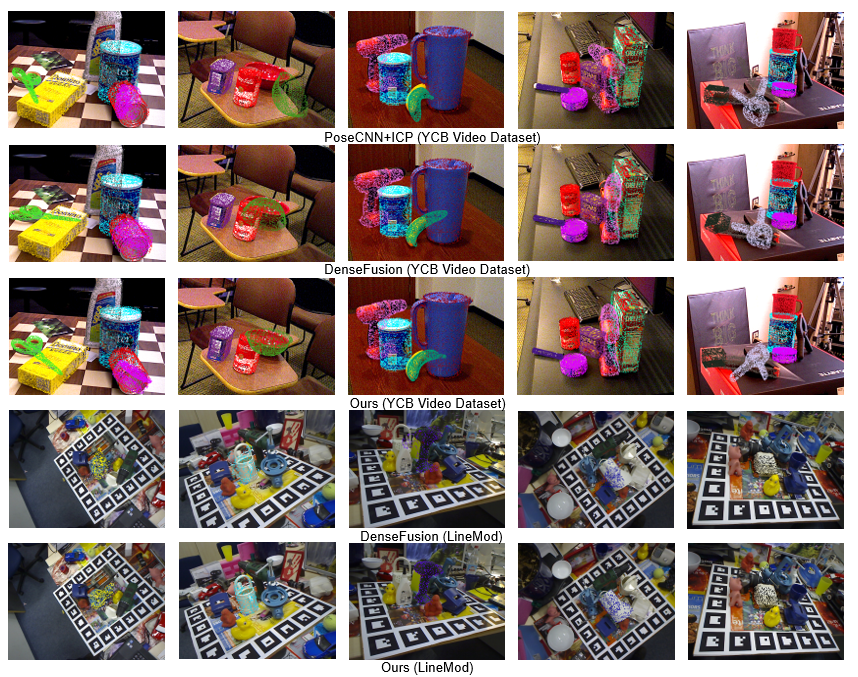}
    \caption{This figure shows the visualization of the result of estimating the pose for the YCB Video Dataset and the Linemod dataset of the latest models and our network. Here, it shows that the pose is accurately estimated even in situations with occlusion or textureless objects that were not well estimated in the existing network.}
    \label{fig:5}
\end{figure*}

TABLE~\ref{table:4} shows the results of a recent comparison of the assessment of 13 objects in the SOTA method and Linemod Dataset. In without the refinement process, The performance was 6\% improvement over the performance of \cite{wang2019densefusion}. In particular, about eight objects show significant performance improvements. And, it can be seen that the accuracy of the pose after the refinement process is higher than that of the existing methods. In addition, we can see that the feature attention is 3.2\% higher than the existing \cite{wang2019densefusion} by using Attention.
However, in objects such as ape and duck, performance is higher than the current most performing W-pose. These objects are smaller than other objects, and we can see that our network performs well for objects that are a little smaller. However, it can be seen that the performance is insufficient for objects such as lamps and phones. This shows that they are not good at objects larger than w-posenet. fig.~\ref{fig:5} shows the visualized results of our model. The above results show accurately estimating poses that were not found in existing networks.

\begin{figure}
     \centering
     \begin{subfigure}[b]{0.45\linewidth}
         \centering
         \includegraphics[scale=0.4]{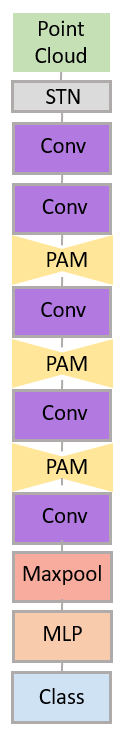}
         \caption{PointNet+PAM}
         \label{Pointnet+PAM}
     \end{subfigure}
     \centering
     \begin{subfigure}[b]{0.45\linewidth}
         \centering
         \includegraphics[scale=0.4]{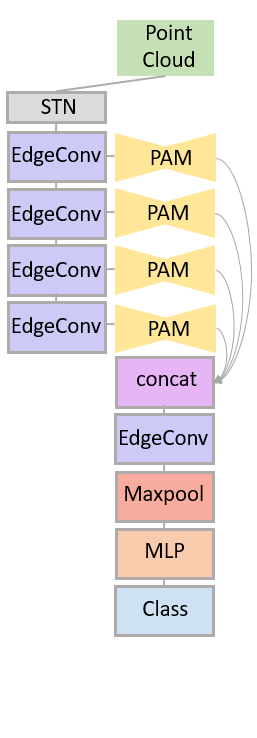}
         \caption{DGCNN+PAM}
         \label{dgcnn+pam}
     \end{subfigure}
     \caption{Architecture that applied PAM to PointNet and DGCNN.By adding PAM in the middle of the layer, we could see a great effect on performance improvement.}
     \label{fig:4}
\end{figure}

\subsection{Application to Classification Task}
We applied the point attention module we applied to another point cloud task, classification. fig.~\ref{fig:4} shows the architecture in which PAM is applied to the middle layer of the Convolution of PointNet and DGCNN models. TABLE~\ref{table:5} shows the results of testing by applying the PAM applied structure to Pointnet and DGCNN to the Modelnet40 classification benchmark dataset. Pointnet structure performance increased from 89.2\% to 91.1\%, and DGCNN structure performance increased from 92.9\% to 93.2\%.It can be seen that it shows a big improvement in performance similar to the network forming the SOTA. This confirmed that our point attention structure can be applied to classification tasks.
\begin{table}
\caption{This table is the result of applying the point-wise attention module to classification. As a result of training on the Modelnet40 dataset and evaluating it, we could see a noticeable performance improvement.}
\label{claasification}
\begin{center}
\begin{tabular}{|c||c||c||c|}
\hline
  Method& Input & \#points & Accuracy(\%)\\ 
\hline\hline
 ECC\cite{simonovsky2017dynamic} & xyz & 1k & 87.4\\
 PointNet\cite{qi2017pointnet}& xyz &1k & 89.2  \\ 
 Kd-Net\cite{klokov2017escape}(depth=10)& xyz &1k & 90.6  \\ 
 PointNet++ \cite{qi2017pointnet++}& xyz &1k & 90.7  \\ 
 KCNet \cite{shen2018mining}& xyz &1k & 91.0  \\ 
 MRTNet\cite{gadelha2018multiresolution}& xyz &1k & 91.2  \\ 
 DGCNN \cite{wang2019dynamic}& xyz &1k & 92.9  \\ 
 SO-Net \cite{li2018so}& xyz &1k & 90.9  \\ 
 KPConv rigid\cite{thomas2019kpconv}& xyz &1k & 92.9  \\ 
 PointNet++ \cite{qi2017pointnet++}& xyz,normal &5k & 91.9  \\ 
 SO-Net\cite{li2018so}& xyz,normal &5k & 93.4  \\ 
\hline\hline
 PointNet+PAM& xyz &1k & 91.1  \\ 
 DGCNN + PAM & xyz & 1k & 93.2\\
\hline
\end{tabular}
\end{center}

\label{table:5}
\end{table}

\section{CONCLUSIONS}
In this paper, a new point-wise attention Module was proposed to estimate the 6D pose when RGB-D images were given. In our approach, attention map was created by encoding each important information using GAPath and CAP. A robust and discriminative feature representation was obtained using the PAM, and as a result, SOTA was achieved in the LineMod and YCB Video Dataset. Additionally, as a result of applying our point-wise Attention Module to the classification task models PointNet and DGCNN, we can see a noticeable performance improvement.



\section*{ACKNOWLEDGMENT}

This work was partly supported by Institute for Information \& communications Technology Promotion(IITP) grant funded by the Korea government(MSIT) (No.2018-0-00677, Development of Robot Hand Manipulation Intelligence to Learn Methods and Procedures for Handling Various Objects with Tactile Robot Hands), and by the Industrial Strategic Technology Development Program(10077538, Development of manipulation technologies in social contexts for human-care service robots) funded by the Ministry of Trade, Industry \& Energy(MOTIE, Korea).


\bibliographystyle{IEEEtran}
\bibliography{IEEEabrv,reference}
\end{document}